\documentclass{esannV2}
\usepackage[dvips]{graphicx}
\usepackage[latin1]{inputenc}
\usepackage{amssymb,amsmath,array}
\usepackage{cite}

\usepackage{listings}
\usepackage{ifxetex}
\usepackage{fancyhdr}
\usepackage{hyperref}
\usepackage{graphicx}
\usepackage{wrapfig}
\usepackage{amsmath}
\usepackage{amsfonts}
\usepackage{amssymb}
\usepackage{amsthm}
\usepackage{listings}
\usepackage{setspace}
\usepackage{wrapfig}
\usepackage{tablefootnote}
\usepackage{multirow}
\usepackage{booktabs}
\usepackage[section]{placeins}
\usepackage{color}
\usepackage{amsfonts}
\usepackage{ulem}
\usepackage{algorithm}
\usepackage{algorithmicx}
\usepackage{algpseudocode}
\usepackage{graphicx}
\usepackage{float}
\usepackage{dsfont}
\usepackage{cleveref}

\allowdisplaybreaks

\newcommand{\mat}[1]{\mathbf{#1}}
\newcommand{\set}[1]{\mathcal{#1}}
\newcommand{\pnorm}[1]{\lVert{#1}\rVert}

\DeclareMathOperator*{\loss}{{\ell}}
\DeclareMathOperator*{\dist}{{d}}
\DeclareMathOperator*{\mad}{{MAD}}
\DeclareMathOperator*{\median}{{median}}

\DeclareMathOperator*{\diag}{diag}

\newcommand{\I}{\mat{\mathbb{I}}}

\newcommand{\RN}{\mathbb{R}}

\DeclareMathOperator*{\regularization}{{\theta}}
\DeclareMathOperator*{\prototype}{{p}}
\newcommand{\x}{\ensuremath{\vec{x}}}

\newcommand{\xcf}{\ensuremath{\vec{x}'}}
\newcommand{\ycf}{\ensuremath{y^{c}}}
\newcommand{\classifier}{\ensuremath{h}}
\newcommand{\protolabel}{\ensuremath{o}}
\DeclareMathOperator*{\distmat}{{\mat{\Omega}}}
\DeclareMathOperator*{\distmatfactor}{{\mat{\tilde{\Omega}}}}

\newtheorem{definition}{Definition}

\begin{document}
\title{Efficient computation of counterfactual explanations of LVQ models}

\author{Andr\'e Artelt\footnote{corresponding author: \href{mailto:aartelt@techfak.uni-bielefeld.de}{aartelt@techfak.uni-bielefeld.de}}\; and Barbara Hammer
%
\thanks{We gratefully acknowledge funding from the VW-Foundation for the project \textit{IMPACT} funded in the frame of the funding line \textit{AI and its Implications for Future Society}.}
%
\vspace{.3cm}\\
%
CITEC - Cognitive Interaction Technology \\
Bielefeld University - Faculty of Technology \\
Inspiration 1, 33619 Bielefeld - Germany
}

\maketitle

\begin{abstract}
The increasing use of machine learning in practice and  legal regulations like EU's GDPR cause the necessity to be able to explain the prediction and behavior of machine learning models. A prominent example of particularly intuitive explanations of AI models in the context of decision making are counterfactual explanations. Yet,  it is still an open research problem how to efficiently compute counterfactual explanations for many models.
We investigate how to efficiently compute counterfactual explanations for an important class of models, prototype-based classifiers such as learning vector quantization models. In particular, we derive specific convex and non-convex programs depending on the used metric.
\end{abstract}

\section{Introduction}
Due to the recent advances in machine learning (ML), ML models are being more and more used in practice and applied to real-world scenarios for decision making~\cite{Goel2016PrecinctOP,creditscoresunfair,creditriskml}.  Essential demands for user acceptance as well as  legal regulations like the EU's "General Data Protection Right" (GDPR)~\cite{gdpr}, that contains a "right to an explanation", make it indispensable to explain the output and behavior of ML models in a comprehensible way.

As a consequence, many research approaches focused on the question how to realize  explainability and transparency in machine learning in recent years~\cite{explainingexplanations, explainingblackboxmodelssurvey, surveyxai, explainableartificialintelligence}. There exist diverse methods for explaining  ML models~\cite{explainingblackboxmodelssurvey,molnar2019}.
One family of methods are model-agnostic methods~\cite{explainingblackboxmodelssurvey,modelagnosticinterpretability}. Model-agnostic methods are flexible in the sense that they are not tailored to a particular model or representation. This makes model-agnostic methods (in theory) applicable to many different types of ML models. In particular, "truly" model-agnostic methods do not need access to the training data or model internals. It is sufficient to have an interface for passing data points to the model and obtaining the output/predictions of the model - the underlying model is viewed as a black-box.

Examples of model-agnostic methods are feature interaction methods~\cite{featureinteraction}, feature importance methods~\cite{featureimportance}, partial dependency plots~\cite{partialdependenceplots} and local methods that approximates the model locally by an explainable model (e.g. a decision tree)~\cite{lime2016,decisiontreecounterfactual}. These methods explain the models by using features as vocabulary.

A different class of model-agnostic explanations are example-based explanations where a prediction or behavior is explained by a (set of) data points~\cite{casebasedreasoning}. Instances of example-based explanations are prototypes \& criticisms~\cite{prototypescriticism} and influential instances~\cite{influentialinstances}. Another instance of example-based explanations are counterfactual explanations~\cite{counterfactualwachter}. A counterfactual explanation is a change of the original input that leads to a different (specific) prediction/behavior of the ML model - \textit{what has to be different in order to change the prediction of the model?} Such an explanation is considered to be fairly intuitive, human-friendly and useful because it tells people what to do in order to achieve a desired outcome~\cite{counterfactualwachter,molnar2019}.

Existing counterfactual explanations are mostly model agnostic methods, which are universally applicable but computationally expensive\cite{decisiontreecounterfactual,counterfactualsgeneticalgorithm,growingsphere}. In this work, we address the question how to efficiently compute counterfactuals for a popular class of models by referring to its specific structure.

Prototype based models such as learning vector quantization (LVQ) represent data by a set of representative samples~\cite{lvqreview}. LVQ models can be combined with metric learning and thereby increase the effectiveness of the model in case of few prototypes~\cite{gmlvq,mrslvq}.
Furthermore, LVQ models can be used in many settings like life-long learning~\cite{lifelonglvq}.

Here, we will consider the question how to efficiently compute counterfactual explanations of prototype-based classifiers, in particular LVQ models.
By exploiting the special structure of such models, we are able to
\begin{enumerate}
\item propose model- and regularization-dependent methods for efficiently computing counterfactual explanations of LVQ models,
\item and empirically demonstrate the efficiency of the modeling as regards speed-up as well as required amount of change in comparison to standard techniques.
\end{enumerate}
Further, the framework enables a straightforward incorporation of domain knowledge, which can be phrased as additional constraints. Such domain knowledge could be used for generating more plausible and feasible counterfactual explanations.\\\\
The remainder of this paper is structured as follows: First, we briefly review counterfactual explanations (section~\ref{sec:counterfactualexplanations}) and learning vector quantization models (section~\ref{sec:lvq}). Then, in section~\ref{sec:cf_lvq} we introduce our convex and non-convex programming framework for efficiently computing counterfactual explanations of different types of LVQ models - note that all derivations can be found in the appendix (section~\ref{sec:appendix}). Furthermore, we empirically evaluate the efficiency of our methods. Finally, section~\ref{sec:conclusion} summarizes the results of this paper.

\section{Counterfactual explanations}\label{sec:counterfactualexplanations}
Counterfactual explanations~\cite{counterfactualwachter} (often just called counterfactuals) are an instance of example-based explanations~\cite{casebasedreasoning}. Other instances of example-based explanations~\cite{molnar2019} are influential instances~\cite{influentialinstances} and prototypes \& criticisms~\cite{prototypescriticism}.

A counterfactual states a change to some features of a given input such that the resulting data point (called counterfactual) has a different (specified) prediction than the original input. The rational is considered to be intuitive, human-friendly and useful because it tells practitioners which minimum  changes can lead to a desired outcome~\cite{counterfactualwachter,molnar2019}.

A classical use case of counterfactual explanations is loan application~\cite{creditriskml,molnar2019}:
\textit{Imagine you applied for a credit at a bank. Unfortunately, the bank rejects your application. Now, you would like to know why. In particular, you would like to know what would have to be different so that your application would have been accepted. A possible explanation might be that you would have been accepted if you would earn 500\$ more per month and if you would not have a second credit card.}

Although counterfactuals constitute very intuitive explanation mechanisms, there do exist a couple of problems.

One problem is that there often exist more than one counterfactual - this is called \textit{Rashomon effect}~\cite{molnar2019}. If there are more than one possible explanation (counterfactual), it is not clear which one should be selected.


In our opinion, the main issue is feasibility and plausibility - in the sense that the counterfactual data point should be valid and plausible in the given domain. There exist first approaches like~\cite{counterfactualguidedbyprototypes, face} but in general it is still an open research question how to efficiently compute feasible and plausible counterfactuals.

An alternative - but very similar in the spirit - to counterfactuals~\cite{counterfactualwachter} is the Growing Spheres method~\cite{growingsphere}. However, this method suffers from the curse of dimensionality because it has to draw samples from the input space, which can become difficult if the input space is high-dimensional.

\begin{definition}[Counterfactual explanation]
Assume a prediction function $\classifier$ is given. Computing a counterfactual $\xcf \in \RN^d$ for a given input $\x \in \RN^d$ is phrased as an optimization problem~\cite{counterfactualwachter}:
\begin{equation}\label{eq:counterfactualoptproblem}
\underset{\xcf \,\in\, \RN^d}{\arg\min}\; \loss\big(\classifier(\xcf), \ycf\big) + C \cdot \regularization(\xcf, \x)
\end{equation}
where $\loss(\cdot)$ denotes the loss function,  $\ycf$ the requested prediction, and  $\regularization(\cdot)$  a penalty term for deviations of $\xcf$ from the original input $\x$. $C>0$ denotes the regularization strength.
\end{definition}

Two common regularizations are the weighted Manhattan distance and the generalized L2 distance.

The weighted Manhattan distance is defined as
\begin{equation}\label{eq:weighted_l1}
\regularization(\xcf, \x) = \sum_j \alpha_j \cdot |(\x)_j - (\xcf)_j|
\end{equation}
where $\alpha_j > 0$ denote the feature wise weights. A popular choice~\cite{counterfactualwachter} for $\alpha_j$ is the inverse median absolute deviation of the $j$-th feature median in the training data set $\set{D}$:
\begin{equation}
\begin{split}
& \alpha_j = \frac{1}{{\mad}_j} \\
& \text{where } {\mad}_j = \underset{\x \,\in\, \set{D}}{\median}\left(\left|(\x)_j - \underset{\x \,\in\, \set{D}}{\median}\big((\x)_j\big)\right|\right)
\end{split}
\end{equation}
The benefit of this choice is that it takes the (potentially) different variability of the features into account.  However, because we need access to the training data set $\set{D}$, this regularization is not a truly model-agnostic method - it is not usable if we only have access to a prediction interface of a black-box model.

The generalized L2 distance\footnote{also called Mahalanobis distance} is defined as
\begin{equation}\label{eq:general_l2}
\regularization(\xcf, \x) = \pnorm{\x - \xcf}_{\distmat}^2 = (\x - \xcf)^\top\distmat(\x - \xcf)
\end{equation}
where $\distmat$ denotes a symmetric positive semi-definite (s.psd) matrix. Note that the L2 distance can be recovered by setting $\distmat=\I$. The generalized L2 distance can be interpreted as the Euclidean distance in a linearly distorted space.

Depending on the model and the choice of $\loss(\cdot)$ and $\regularization(\cdot)$, the final optimization problem might be differentiable or not. If it is differentiable, we can use a gradient-based optimization algorithm like conjugate gradients, gradient descent or (L-)BFGS. Otherwise, we have to use a black-box optimization algorithm like Downhill-Simplex (also called Nelder-Mead) or Powell. However, using a model and regularization specific optimization technique has the best chances to enable a simplification and computationally efficient solution. Unfortunately, there exit model specific optimizers for counterfactual explanations for few ML models only.

\section{Learning vector quantization}\label{sec:lvq}
In learning vector quantization (LVQ) models~\cite{lvqreview} we compute a set of labeled prototypes $\{(\vec{\prototype}_i, \protolabel_i)\}$ from a training data set of labeled real-valued vectors - we refer to the $i$-th prototype as $\vec{\prototype}_i$ and the corresponding label as $\protolabel_i$.
A new data point is classified according to the winner-takes-it-all scheme:
\begin{equation}\label{eq:lvq_predict}
\begin{split}
& \classifier(\x) = \protolabel_i \quad \text{s.t. } \vec{\prototype}_i = \underset{\vec{\prototype}_j}{\arg\min}\;{\dist}(\x, \vec{\prototype}_j)
\end{split}
\end{equation}
where $\dist(\cdot)$ denotes a function for computing the distance between a data point and a prototype - in vanillas LVQ, this is independent chosen globally as the squared Euclidean distance:
\begin{equation}\label{eq:distfunction}
\dist(\x, \vec{\prototype}) = (\x - \vec{\prototype})^\top \I (\x - \vec{\prototype})
\end{equation}
There exist matrix-LVQ models like GMLVQ, LGMLVQ~\cite{gmlvq}, MRSLVQ and LMRSLVQ~\cite{mrslvq} that learn a custom (class or prototype specific) distance matrix $\distmat_{\prototype}$ that is used instead of the identity $\I$ when computing the distance between a data point and a prototype. This gives rise to the generalized L2 distance Eq.~\eqref{eq:general_l2}:
\begin{equation}\label{eq:lvq_generaldist}
\dist(\x, \vec{\prototype}) = (\x - \vec{\prototype})^\top{\distmat}_{\prototype}(\x - \vec{\prototype})
\end{equation}
Because ${\distmat}_{\prototype}$ must be a s.psd matrix, instead of learning ${\distmat}_{\prototype}$ directly, these LVQ variants learn a matrix $\distmatfactor_{\prototype}$ and compute the final distance matrix as:
\begin{equation}
{\distmat}_{\prototype}={\distmatfactor}_{\prototype}^\top{\distmatfactor}_{\prototype}
\end{equation}
By this, we can guarantee that the matrix $\distmat_{\prototype}$ is s.psd, whereas the model only has to learn an arbitrary matrix ${\distmatfactor}_{\prototype}$ - which is much easier than making some restriction on the definiteness. Training usually takes place by optimizing suitable cost functions as regards prototype locations and metric parameters. For counterfactual reasoning, the specific training method~\cite{gmlvq,mrslvq} is irrelevant and we refer to the final model only.

\section{Counterfactual explanations of LVQ models}\label{sec:cf_lvq}
We aim for an efficient explicit formulation how to find counterfactuals, for diverse  LVQ models.

\subsection{General approach}
Because a LVQ model assigns the label of the nearest prototype to a given input, we know that the nearest prototype of a counterfactual $\xcf$ must be a prototype $\vec{\prototype}_i$ with $\protolabel_i=\ycf$. In order to compute a counterfactual $\xcf$ of a given input $\x$, it is sufficient to solve the following optimization problem for each prototype $\vec{\prototype}_i$ with $\protolabel_i=\ycf$ and select the counterfactual $\xcf$ yielding the smallest value of $\regularization(\xcf, \x)$:
\begin{subequations}
\label{eq:cflvq_general}
 \begin{align}
  &\underset{\xcf \,\in\, \RN^d}{\arg\min}\;\regularization(\xcf, \x) \label{eq:cflvq_general:regularization} \\
  & \quad \text{s.t. } \dist(\xcf, \vec{\prototype}_i) + \epsilon \leq \dist(\xcf, \vec{\prototype}_j) \quad \forall\, \vec{\prototype}_j\in\set{P}(\ycf) \label{eq:cflvq_general:constraints}
 \end{align}
\end{subequations}
where $\set{P}(\ycf)$ denotes the set of all prototypes \emph{not} labeled as $\ycf$ and $\epsilon>0$ is a small value preventing that the counterfactual lies exactly on the decision boundary. Note that the feasible region of Eq.~\eqref{eq:cflvq_general} is always non-empty - the prototype $\vec{\prototype}_i$ is always a feasible solution. Furthermore, in contrast to Eq.~\eqref{eq:counterfactualoptproblem}, the formulation does not include hyperparameters.

The pseudocode for computing a counterfactual of a LVQ model is described in Algorithm~\ref{algo:cflvq}.
\begin{algorithm}[!tb]
\caption{Computing a counterfactual of a LVQ model}\label{algo:cflvq}
\textbf{Input:} Original input $\x$, requested prediction $\ycf$ of the counterfactual, the LVQ model\\
\textbf{Output:} Counterfactual $\xcf$
\begin{algorithmic}[1]
 \State $\xcf = \vec{0}$ \Comment{Initialize dummy solution}
 \State $z = \infty$
 \For{$\vec{\prototype}_i$ with $\protolabel_i=\ycf$} \Comment{Try each prototype with the correct label}
 	\State Solving Eq.~\eqref{eq:cflvq_general} yields a counterfactual $\xcf_{*}$
 	\If{$\regularization(\xcf_{*}, \x) < z$} \Comment{Keep this counterfactual if it deviates less from the original input then the currently "best" counterfactual}
 		\State $z=\regularization(\xcf_{*}, \x)$
 		\State $\xcf = \xcf_{*}$
 	\EndIf
 \EndFor
\end{algorithmic}
\end{algorithm}
Note that the \texttt{for} loop in line 3 of Algorithm~\ref{algo:cflvq} can be easily parallelized.

For the weighted Manhattan distance as a regularization $\regularization(\cdot)$, the objective Eq.~\eqref{eq:cflvq_general:regularization} becomes linear in $\xcf$, where
$\mat{\Upsilon}$ is the diagonal matrix with entries $\alpha_j$ and $\vec{\beta}$ is an auxiliary variable that can be discarded afterwards:
\begin{equation}\label{eq:objective:l1}
\begin{split}
& \underset{\xcf, \vec{\beta} \,\in\, \RN^d}{\min}\;\vec{1}^\top\vec{\beta} \\
& \text{s.t.} \;\mat{\Upsilon}\xcf - \mat{\Upsilon}\x \leq \vec{\beta}, \quad -\mat{\Upsilon}\xcf + \mat{\Upsilon}\x \leq \vec{\beta}, \quad\vec{\beta} \geq \vec{0}
\end{split}
\end{equation}
For the Euclidean distance Eq.~\eqref{eq:general_l2} as  regularization $\regularization(\cdot)$, the objective Eq.~\eqref{eq:cflvq_general:regularization} can be written in a convex quadratic form:
\begin{equation}\label{eq:objective:l2}
\underset{\xcf \,\in\, \RN^d}{\min}\;\frac{1}{2}\xcf^\top\xcf - \xcf^\top\x
\end{equation}
\mbox{}\\
In the subsequel  we explore Eq.~\eqref{eq:cflvq_general} for different regularizations $\regularization(\cdot)$ and LVQ models, and investigate how to solve it efficiently. Note that for the purpose of better readability and due to space constraints, we put all derivations in the appendix (section~\ref{sec:appendix}).

\subsection{(Generalized matrix) LVQ}
When using (generalized matrix) LVQ models - no class or prototype specific distance matrices - the optimization problem Eq.~\eqref{eq:cflvq_general} becomes a linear program (LP) when using the weighted Manhattan distance as a regularizer, and a convex quadratic program (QP) problem when using the Euclidean distance. Both programs can be solved efficiently and (up to equivalence) uniquely~\cite{Boyd2004}.
More precisely, 
the constraints Eq.~\eqref{eq:cflvq_general:constraints} can be written as a set of linear inequality constraints:
\begin{equation}
\xcf^\top\vec{q}_{ij} + r_{ij} + \epsilon\leq 0 \quad \forall\, \vec{\prototype}_j\in\set{P}(\ycf) 
\end{equation}
where
\begin{equation}
\vec{q}_{ij} = \distmat(\vec{\prototype}_{j} - \vec{\prototype}_i) \quad\quad r_{ij} =  \frac{1}{2}\left(\vec{\prototype}_{i}^\top\distmat\vec{\prototype}_{i} - \vec{\prototype}_{j}^\top\distmat\vec{\prototype}_{j}\right)
\end{equation}
We set $\distmat=\I$ if the LVQ model uses the Euclidean distance.

Another benefit is that we can easily add additional convex quadratic constraints like box constraints, restricting the value range of linear interaction of features or freezing some features - these features are not allowed to be different in the counterfactual. We can use such additional constraints for computing a more plausible and feasible counterfactual.

\subsection{(Localized generalized matrix) LVQ}
In case of local matrix-LVQ models that learn a class or prototype specific distance matrix $\distmat_{\prototype}$, the optimization problem Eq.~\eqref{eq:cflvq_general} becomes a quadratically constrained quadratic program (QCQP) for Manhattan and Euclidean regularization, since
the constraints Eq.~\eqref{eq:cflvq_general:constraints} become a set of quadratic constraints:
\begin{equation}\label{eq:lgmlvq:constraints}
\frac{1}{2}\xcf^\top\mat{Q}_{ij}\xcf + \xcf^\top\vec{q}_{ij} + r_{ij} + \epsilon \leq 0 \quad \forall\, \vec{\prototype}_j\in\set{P}(\ycf)
\end{equation}
where
\begin{equation}
\mat{Q}_{ij} = {\distmat}_{i} - {\distmat}_{j}
\end{equation}
Unfortunately, we can not make any statement about the definiteness of $\mat{Q}_{ij}$. Because $\mat{Q}_{ij}$ is the difference of two s.psd matrices, all we know is that it is a symmetric matrix. Therefore, Eq.~\eqref{eq:lgmlvq:constraints} yields a non-convex QCQP and solving a non-convex QCQP is known to be NP-hard~\cite{park2017general}. However, there exist methods like the Suggest-Improve framework~\cite{park2017general} that can approximately solve a non-convex QCQP very efficiently - more details on how to apply this to Eq.~\eqref{eq:lgmlvq:constraints} can be found in the appendix (section~\ref{sec:appendix}).

\subsection{Experiments}
We empirically confirm the efficiency of our proposed methods in comparison to black-box mechanisms by means of the following experiments:
We use GLVQ, GMLVQ and LGMLVQ models with 3 prototypes per class for the "Breast Cancer Wisconsin (Diagnostic) Data Set"~\cite{breastcancer}, the "Optical Recognition of Handwritten Digits Data Set"~\cite{ocr} and the "Ames Housing dataset"~\cite{housepricedata}. Thereby, we use PCA-preprocessing
\footnote{This is for the purpose of better stability and better semantic meaning, since in the original domain already a small perturbation is sufficient for changing the class, since adversarial attacks exist even for linear functions in high dimensions if feature correlations are neglected. Since PCA can be approximately inverted, counterfactuals in PCA space can be lifted to the original data space.} to reduce the dimensionality of the digit data set to $10$ and of the breast cancer data set to $5$. We standardize the house data set and turned it into a binary classification problem\footnote{In addition, we select the following features: TotalBsmt, 1stFlr, 2ndFlr, GrLivA, WoodDeck, OpenP, 3SsnP, ScreenP and PoolA - When computing counterfactuals, we fix the last five features.} by setting the target to $1$ if the price is greater or equal to 160k\$ and $0$ otherwise.
We use the Suggest-Improve framework~\cite{park2017general} for  solving the non-convex QCQPs, where we pick the target prototype as initial  solution in the Suggest-step and we use the penalty convex-concave procedure (CCP) in the Improve-step~\cite{park2017general}.
For comparison, we use the optimizer for computing counterfactual explanations of LVQ models as implemented in ceml~\cite{ceml} - where the distance to the nearest prototype with the requested label $\ycf$ is minimized by  Downhill-Simplex search or CMA-ES.

We report results for the Manhattan distance as regularizer - we used the Manhattan distance for enforcing a sparse solution.
For each possible combination of model, data set and method,  a 4-fold cross validation is conducted and the mean  distance is reported.
The results are listed in Table~\ref{table:experimentresults1}. In all cases, our method yields counterfactuals that are closer to the original data point than the one found by minimizing the original cost function Eq.~\eqref{eq:counterfactualoptproblem} with Downhill-Simplex search (DS) or CMA-ES.
In addition, our method is between $1.5$ and $158.0$ faster in comparison to  DS/CMA-ES method. Furthermore, Downhill-Simplex and CMA-ES did not always find a counterfactual when dealing with LGMLVQ models.
We would like to remark that our formulation can easily be extended by linear/quadratic constraints which can incorporate prior knowledge such as a maximum possible change of specific input features - see Table~\ref{table:pausiblecounterfactuals} for an example. Such extensions do not change the form of the optimization problem hence its complexity.

\begin{table}[t]
\centering
\footnotesize
\begin{tabular}{|c||c|c|c||c|c|c||c|c|c|}
 \hline
 \multicolumn{1}{|c||}{\textit{Data set}} & \multicolumn{3}{|c||}{Breast cancer} & \multicolumn{3}{|c||}{Handwritten digits} & \multicolumn{3}{|c|}{House prices}  \\
  \hline
 \textit{ Method} & DS & CMA & Ours & DS & CMA & Ours & DS & CMA & Ours \\
 \hline
 GLVQ   & 3.26 & 3.28 & \textbf{1.96} & 6.51 & 6.53 & \textbf{3.99} & 3.81 & 3.85 & \textbf{3.32} \\
 GMLVQ  & 2.71 & 6.49 & \textbf{2.46} & 21.34 & 11.63 & \textbf{4.40} & 5.06 & 8.63 & \textbf{3.78} \\
LGMLVQ & {\it 2.00}
 & {\it 1.61}
 & \textbf{1.57} & {\it 8.12}
 & {\it 7.88}
 & \textbf{7.53} & {\it 12.74}
 & {\it 12.59}
 & \textbf{8.20} \\
 \hline
\end{tabular}
\caption{Mean \emph{Manhattan distance} between the counterfactual and the original data point - best (smallest) values are \textbf{highlighted}.
For LGMLVQ with DS or CMA-ES (marked italic), in  5\% to 60\% of the cases no solution was found.}
\label{table:experimentresults1}
\end{table}
\begin{table}[htb]
\centering
\footnotesize
\begin{tabular}{|c||c|c|c|c|c|}
 \hline
 \multicolumn{1}{|c||}{\textit{Data point}} & \multicolumn{1}{|c|}{TotalBsmt} & \multicolumn{1}{|c|}{1stFlr} & \multicolumn{1}{|c|}{2ndFlr} & \multicolumn{1}{|c|}{GrLivA} 
 & \multicolumn{1}{|c|}{Label} \\
  \hline
 \textit{Original} & 0 & 1120 & 468 & 1588 
 & 1 \\
 \hline
 \textit{Counterfactual} & 0 & 366 & 1824 & 2225 
 & 0 \\
 \textit{Constrained Counterfactual}
 & 373 & 1454 & 1454 & 3125 
 & 0 \\
 \hline
 \end{tabular}
 \caption{House prices, we obtain a "plausible" counterfactual by adding constrains, here the constraint "2ndFlr $\leq$ 1stFlr" is added.}
\label{table:pausiblecounterfactuals}
\end{table}
The implementation of our proposed method for computing counterfactual explanations is available online\footnote{\url{https://github.com/andreArtelt/efficient_computation_counterfactuals_lvq}}. All experiments were implemented in Python 3.6~\cite{van1995python} using the packages cvxpy~\cite{cvxpy} cvx-qcqp~\cite{qcqp}, sklearn-lvq~\cite{sklearn_lvq}, numpy~\cite{numpy}, scipy~\cite{scipy}, scikit-learn~\cite{scikit-learn} and ceml~\cite{ceml}.

\section{Conclusion}\label{sec:conclusion}
We proposed, and empirically evaluated, model- and regularization-dependent convex and non-convex programs for efficiently computing counterfactual explanations of LVQ models. We found that in many cases we get either a set of linear or convex quadratic programs which both can be solved efficiently. Only in the case of localized matrix-LVQ models we have to solve a set of non-convex quadratically constrained quadratic programs - we found that they can be efficiently approximately solved by using the Suggest-Improve framework.

\section{Appendix}\label{sec:appendix}
\subsection{Minimizing the Euclidean distance}
First, we expand the Euclidean distance Eq.~\eqref{eq:general_l2}:
\begin{equation}
\begin{split}
\pnorm{\xcf - \x}_2^2 &= (\xcf - \x)^\top(\xcf - \x) \\
&= \xcf^\top\xcf - \xcf^\top\x - \x^\top\xcf + \x^\top\x \\
&= \xcf^\top\xcf - 2\x^\top\xcf + \x^\top\x \\
\end{split}
\end{equation}
Next, we note that that we can drop the constant $\x^\top\x$ when optimizing with respect to $\xcf$:
\begin{equation}
\begin{split}
& \underset{\xcf \,\in\, \RN^d}{\min}\; \pnorm{\xcf - \x}_2^2 \\
& \Leftrightarrow \\
& \underset{\xcf \,\in\, \RN^d}{\min}\; \frac{1}{2}\xcf^\top\xcf - \x^\top\xcf
\end{split}
\end{equation}

\subsection{Minimizing the weighted Manhattan distance}
First, we transform the problem of minimizing the weighted Manhattan distance Eq.~\eqref{eq:weighted_l1} into epigraph form:
\begin{equation}
\begin{split}
& \underset{\xcf \,\in\, \RN^d}{\min}\; \sum_j \alpha_j \cdot |(\xcf)_j - (\x)_j| \\
& \Leftrightarrow \\
& \underset{\xcf \,\in\, \RN^d, \beta\,\in\, \RN}{\min}\; \beta \\
& \quad \text{s.t.} \quad \sum_j \alpha_j \cdot |(\xcf)_j - (\x)_j| \leq \beta \\
& \quad \quad \quad \beta \geq 0
\end{split}
\end{equation}
Next, we separate the dimensions:
\begin{equation}
\begin{split}
& \underset{\xcf \,\in\, \RN^d, \beta\,\in\, \RN}{\min}\; \beta \\
& \quad \text{s.t.} \quad \sum_j \alpha_j \cdot |(\xcf)_j - (\x)_j| \leq \beta \\
& \quad \quad \quad \beta \geq 0 \\
& \Leftrightarrow \\
& \underset{\xcf, \vec{\beta} \,\in\, \RN^d}{\min}\; \sum_j (\vec{\beta})_j \\
& \quad \text{s.t.} \quad \alpha_j \cdot |(\xcf)_j - (\x)_j| \leq (\vec{\beta})_j \quad \forall\,j \\
& \quad \quad \quad (\vec{\beta})_j \geq 0 \quad \forall\,j
\end{split}
\end{equation}
After that, we remove the absolute value function:
\begin{equation}
\begin{split}
& \underset{\xcf, \vec{\beta} \,\in\, \RN^d}{\min}\; \sum_j (\vec{\beta})_j \\
& \quad \text{s.t.} \quad \alpha_j \cdot |(\xcf)_j - (\x)_j| \leq (\vec{\beta})_j \quad \forall\,j \\
& \quad \quad \quad (\vec{\beta})_j \geq 0 \quad \forall\,j \\
& \Leftrightarrow \\
& \underset{\xcf, \vec{\beta} \,\in\, \RN^d}{\min}\; \sum_j (\vec{\beta})_j \\
& \quad \text{s.t.} \quad \alpha_j (\xcf)_j - \alpha_j (\x)_j \leq (\vec{\beta})_j \quad \forall\,j \\
& \quad \quad \quad -\alpha_j (\xcf)_j + \alpha_j (\x)_j \leq (\vec{\beta})_j \quad \forall\,j \\
& \quad \quad \quad (\vec{\beta})_j \geq 0 \quad \forall\,j
\end{split}
\end{equation}
Finally, we rewrite everything in matrix-vector notation:
\begin{equation}
\begin{split}
& \underset{\xcf, \vec{\beta} \,\in\, \RN^d}{\min}\;\vec{1}^\top\vec{\beta} \\
& \text{s.t.} \\
& \mat{\Upsilon}\xcf - \mat{\Upsilon}\x \leq \vec{\beta} \\
& -\mat{\Upsilon}\xcf + \mat{\Upsilon}\x \leq \vec{\beta} \\
& \vec{\beta} \geq \vec{0}
\end{split}
\end{equation}
where
\begin{equation}
\mat{\Upsilon} = \diag(\alpha_j)
\end{equation}

\subsection{Enforcing a specific prototype as the nearest neighbor}
By using the following set of strict inequalities, we can force the prototype $\vec{\prototype}_i$ to be the nearest neighbor of the counterfactual $\xcf$ - which would cause $\xcf$ to be classified as $\protolabel_i$: 
\begin{equation}
\dist(\xcf, \vec{\prototype}_i) < \dist(\xcf, \vec{\prototype}_j) \quad \forall\, \vec{\prototype}_j\in\set{P}(\ycf)
\end{equation}
We consider a fixed pair of $i$ and $j$:
\begin{equation}\label{eq:nearestprototypeconstraint}
\begin{split}
& \dist(\xcf, \vec{\prototype}_i) < \dist(\xcf, \vec{\prototype}_j) \\
& \Leftrightarrow \pnorm{\xcf - \vec{\prototype}_i}_{\distmat_i}^2 < \pnorm{\xcf - \vec{\prototype}_j}_{\distmat_j}^2 \\
& \Leftrightarrow (\xcf - \vec{\prototype}_i)^\top{\distmat}_i(\xcf - \vec{\prototype}_i) < (\xcf - \vec{\prototype}_j)^\top{\distmat}_j(\xcf - \vec{\prototype}_j) \\
& \Leftrightarrow \xcf^\top{\distmat}_i\xcf - 2\xcf^\top{\distmat}_i\vec{\prototype}_i + \vec{\prototype}_i^\top{\distmat}_i\vec{\prototype}_i < \xcf^\top{\distmat}_j\xcf - 2\xcf^\top{\distmat}_j\vec{\prototype}_j + \vec{\prototype}_j^\top{\distmat}_i\vec{\prototype}_j \\
& \Leftrightarrow \xcf^\top{\distmat}_i\xcf - \xcf^\top{\distmat}_j\xcf - 2\xcf^\top{\distmat}_i\vec{\prototype}_i + 2\xcf^\top{\distmat}_j\vec{\prototype}_j + \vec{\prototype}_i^\top{\distmat}_i\vec{\prototype}_i - \vec{\prototype}_j^\top{\distmat}_i\vec{\prototype}_j < 0 \\
& \Leftrightarrow \xcf^\top({\distmat}_i - {\distmat}_j)\xcf + \xcf^\top(-2{\distmat}_i\vec{\prototype}_i + 2{\distmat}_j\vec{\prototype}_j) + (\vec{\prototype}_i^\top{\distmat}_i\vec{\prototype}_i - \vec{\prototype}_j^\top{\distmat}_i\vec{\prototype}_j) < 0 \\
& \Leftrightarrow \frac{1}{2}\xcf^\top({\distmat}_i - {\distmat}_j)\xcf + \xcf^\top({\distmat}_j\vec{\prototype}_j - {\distmat}_i\vec{\prototype}_i) + \frac{1}{2}(\vec{\prototype}_i^\top{\distmat}_i\vec{\prototype}_i - \vec{\prototype}_j^\top{\distmat}_i\vec{\prototype}_j) < 0 \\
& \Leftrightarrow \frac{1}{2}\xcf^\top\mat{Q}_{ij}\xcf + \xcf^\top\vec{q}_{ij} + r_{ij} < 0
\end{split}
\end{equation}
where
\begin{equation}
\mat{Q}_{ij} = {\distmat}_{i} - {\distmat}_{j}
\end{equation}
\begin{equation}
\vec{q}_{ij} = {\distmat}_j\vec{\prototype}_{j} - {\distmat}_i\vec{\prototype}_{i}
\end{equation}
\begin{equation}
r_{ij} = \frac{1}{2}\left(\vec{\prototype}_{i}\top{\distmat}_{i}\vec{\prototype}_{i} - \vec{\prototype}_{j}\top{\distmat}_{j}\vec{\prototype}_{j}\right)
\end{equation}
\mbox{}\\
If we only have a global distance matrix $\distmat$, we find that $\mat{Q}_{ij}=\mat{0}$ and the inequality Eq.~\eqref{eq:nearestprototypeconstraint} simplifies:
\begin{equation}
\begin{split}
& \dist(\x, \vec{\prototype}_i) < \dist(\x, \vec{\prototype}_j) \\
& \Leftrightarrow \xcf^\top\vec{q}_{ij} + r_{ij} < 0
\end{split}
\end{equation}
where
\begin{equation}
\vec{q}_{ij} = {\distmat}\left(\vec{\prototype}_{j} - \vec{\prototype}_{i}\right)
\end{equation}
\begin{equation}
r_{ij} = \frac{1}{2}\left(\vec{\prototype}_{i}^\top\distmat\vec{\prototype}_{i} - \vec{\prototype}_{j}^\top\distmat\vec{\prototype}_{j}\right)
\end{equation}
\mbox{}\\
If we do not use a custom distance matrix, we have $\distmat=\I$ and Eq.~\eqref{eq:nearestprototypeconstraint} becomes:
\begin{equation}
\begin{split}
& \dist(\x, \vec{\prototype}_i) < \dist(\x, \vec{\prototype}_j) \\
& \Leftrightarrow \xcf^\top\vec{q}_{ij} + r_{ij} < 0
\end{split}
\end{equation}
where
\begin{equation}
\vec{q}_{ij} = \vec{\prototype}_{j} - \vec{\prototype}_{i}
\end{equation}
\begin{equation}
r_{ij} = \frac{1}{2}\left(\vec{\prototype}_{i}^\top\vec{\prototype}_{i} - \vec{\prototype}_{j}^\top\vec{\prototype}_{j}\right)
\end{equation}

\subsection{Approximately solving the non-convex QCQP}
Recall the non-convex quadratic constraint from Eq.~\eqref{eq:nearestprototypeconstraint}:
\begin{equation}\label{eq:nonconvexquadratic}
\frac{1}{2}\xcf^\top\mat{Q}_{ij}\xcf + \xcf^\top\vec{q}_{ij} + r_{ij} + \epsilon \leq 0
\end{equation}
where the matrix $\mat{Q}_{ij}$ is defined as the difference of two s.psd matrices:
\begin{equation}\label{eq:omegadiff}
\mat{Q}_{ij} = {\distmat}_{i} - {\distmat}_{j}
\end{equation}
By making use of Eq.~\eqref{eq:omegadiff}, we can rewrite Eq.~\eqref{eq:nonconvexquadratic} as:
\begin{equation}
\begin{split}
& \frac{1}{2}\xcf^\top{\distmat}_{i}\xcf + \xcf^\top\vec{q}_{ij} + r_{ij} + \epsilon - \frac{1}{2}\xcf^\top{\distmat}_{j}\xcf \leq 0 \\
& \Leftrightarrow f(\xcf) - g(\xcf) \leq 0
\end{split}
\end{equation}
where
\begin{equation}
f(\xcf) = \frac{1}{2}\xcf^\top{\distmat}_{i}\xcf + \xcf^\top\vec{q}_{ij} + r_{ij}  + \epsilon
\end{equation}
\begin{equation}
g(\xcf) = \frac{1}{2}\xcf^\top{\distmat}_{j}\xcf
\end{equation}
Under the assumption that the regularization function $\regularization(\cdot)$ is a convex function\footnote{The weighted Manhattan distance and the Euclidean distance are convex functions!}, we can rewrite a generic version of the non-convex QCQP (induced by Eq.~\eqref{eq:lgmlvq:constraints}) as follows:
\begin{equation}\label{eq:dcp}
\begin{split}
& \underset{\xcf \,\in\, \RN^d}{\min}\;\regularization(\xcf, \x) \\
& \text{s.t.} \\
& f(\xcf) - g(\xcf) \leq 0
\end{split}
\end{equation}
Because ${\distmat}_{i}$ and ${\distmat}_{j}$ are s.psd matrices, we know that $f(\xcf)$ and $g(\xcf)$ are convex functions. Therefore Eq.~\eqref{eq:dcp} is a difference-of-convex program (DCP).

This allows us to use the penalty convex-concave procedure (CCP)~\cite{park2017general} for computing an approximate solution of Eq.~\eqref{eq:dcp} that is equivalent to the original non-convex QCQP induced by Eq.~\eqref{eq:lgmlvq:constraints}. For using the penalty CCP, we need the first order Taylor approximation of $g(\xcf)$ around a current point $\x_k$:
\begin{equation}
\begin{split}
\hat{g}(\xcf)_{\x_k} &= g(\x_k) + (\nabla_{\xcf}g)(\x_k)^\top(\xcf - \x_k) \\
&= \frac{1}{2}\x_k^\top{\distmat}_{j}\x_k + ({\distmat}_{j}\x_k)^\top(\xcf - \x_k) \\
&= ({\distmat}_{j}\x_k)^\top\xcf + \frac{1}{2}\x_k^\top{\distmat}_{j}\x_k -({\distmat}_{j}\x_k)^\top\x_k \\
&= \vec{\rho}_{jk}^\top\xcf + \tilde{r}_{jk}
\end{split}
\end{equation}
where
\begin{equation}
\vec{\rho}_{jk} = {\distmat}_{j}\x_k
\end{equation}
\begin{equation}
\tilde{r}_{jk} = -\frac{1}{2}\x_k^\top{\distmat}_{j}\x_k
\end{equation}


\begin{footnotesize}




\bibliographystyle{unsrt}
\bibliography{bibliography}

\end{footnotesize}


\end{document}